\documentclass[preprint,12pt]{elsarticle}

\usepackage{amssymb}
\usepackage{amsmath}
\setcitestyle{sort&compress}
\usepackage{url}
\usepackage{booktabs}
\usepackage{subcaption}
\usepackage{tcolorbox}

\newtcolorbox{textbox}[1]{
    colback=blue!5!white,      
    colframe=blue!75!black,    
    fonttitle=\bfseries,
    title={#1},                
    sharp corners,             
    boxrule=1pt,
    left=5pt, right=5pt, top=5pt, bottom=5pt
}
\usepackage{graphicx}
\usepackage{cuted}
\usepackage{tikz}
\usepackage[pagewise]{lineno}
\usepackage{multirow}
\usetikzlibrary{positioning, shapes.geometric, arrows.meta, shadows.blur}

\usepackage[hidelinks]{hyperref}
\hypersetup{
  colorlinks=true,     
  linkcolor=black,     
  citecolor=teal,      
  urlcolor=black,      
  filecolor=black,     
  anchorcolor=black    
}

\usepackage{fontawesome5} 

\journal{Nuclear Physics B}

\begin{document}

\begin{frontmatter}



\title{MultiFoodhat: A potential new paradigm for intelligent food quality inspection} 


\author[1]{Yue Hu}
\author[2]{Guohang Zhuang\corref{cor1}}
\affiliation[1]{organization={School of Food Science and Engineering, Central South University of Forestry and Technology},
                addressline={},
                city={Changsha},
                postcode={410004},
                state={Hunan},
                country={China}}

\affiliation[2]{organization={School of Computer and Information, Hefei University of Technology},
                addressline={Shushan District},
                city={Hefei},
                postcode={230009},
                state={Anhui},
                country={China}}

\cortext[cor1]{Corresponding author. E-mail: guohang\_zhuang@hfut.edu.cn}

\begin{abstract}
Food image classification plays a vital role in intelligent food quality inspection, dietary assessment, and automated monitoring. However, most existing supervised models rely heavily on large labeled datasets and exhibit limited generalization to unseen food categories. To overcome these challenges, this study introduces MultiFoodChat, a dialogue-driven multi-agent reasoning framework for zero-shot food recognition. The framework integrates vision–language models (VLMs) and large language models (LLMs) to enable collaborative reasoning through multi-round visual–textual dialogues. An Object Perception Token (OPT) captures fine-grained visual attributes, while an Interactive Reasoning Agent (IRA) dynamically interprets contextual cues to refine predictions. This multi-agent design allows flexible and human-like understanding of complex food scenes without additional training or manual annotations. Experiments on multiple public food datasets demonstrate that MultiFoodChat achieves superior recognition accuracy and interpretability compared with existing unsupervised and few-shot methods, highlighting its potential as a new paradigm for intelligent food quality inspection and analysis.
\end{abstract}





\end{frontmatter}


\section{Introduction}
Food safety and nutrition monitoring are fundamental issues in modern food science. With the globalization of food supply chains and the increasing diversity of dietary habits, there is a growing demand for accurate, efficient, and scalable food recognition technologies. Reliable identification of food items supports multiple applications, including food safety surveillance~\cite{qian2023can,kudashkina2022artificial,pezo2025modeling}, quality control~\cite{hernansanz2026real,chhetri2024applications,yu2025research}, dietary assessment~\cite{lu2020gofoodtm,cofre2025validity,lo2024dietary}, and intelligent nutrition management~\cite{joshi2024artificial,ma2022deep,mao2024deep}. In this context, food image recognition lies at the intersection of food chemistry and computer vision, providing a data-driven approach to protecting public health and enabling deeper chemical and nutritional analysis of complex food systems~\cite{gao2024high,mezgec2017nutrinet,liu2016deepfood,sahoo2019foodai}.

Early studies in food image recognition relied on handcrafted visual features such as color, texture, and shape. For example, Chen et al.~\cite{chen2009pfid} employed RGB color histograms with SVM classifiers, while Lowe et al.~\cite{lowe2004distinctive} introduced SIFT descriptors for local feature representation. Nguyen et al.~\cite{NGUYEN2014242} further integrated texture and structural information to enhance classification. Although these approaches achieved moderate performance under controlled conditions, they were highly sensitive to illumination changes, occlusion, and complex food backgrounds. The advent of deep learning, particularly Convolutional Neural Networks (CNNs), has significantly improved food recognition. CNN-based models such as ResNet~\cite{he2016deep} and Inception~\cite{szegedy2015going} automatically learn hierarchical visual features and have demonstrated superior performance on benchmark datasets like Food101~\cite{bossard2014food}. Nevertheless, CNNs remain dependent on large-scale annotated datasets, and their generalization is limited when encountering novel food categories, regional cuisines, or noisy real-world data.

Recent progress in large-scale pre-trained models, including vision–language models (VLMs) and large language models (LLMs), has enabled training-free object classification via zero-shot learning. Models such as GPT-4o~\cite{brown2020language} show strong reasoning abilities for semantic image understanding. However, most studies use these models separately—focusing either on visual perception~\cite{zhu2022pointclip} or text-based reasoning~\cite{chen2020generative}—while the potential of collaborative multi-agent reasoning remains underexplored. 

Motivated by these opportunities, we propose MultiFoodChat, a zero-shot, multi-agent framework for collaborative reasoning in object classification. Each agent is specialized for distinct reasoning tasks, including visual grounding, semantic analysis, and integrative summarization. Agents independently generate intermediate conclusions and deliberate collectively to reach final decisions. This multi-agent design reduces reliance on labeled data while enhancing adaptability, robustness, and interpretability. Experiments on four benchmark datasets demonstrate that MultiFoodchat achieves accuracy comparable to state-of-the-art supervised models and substantially outperforms existing single-agent or zero-shot baselines.

\begin{figure}
	\centering
	\includegraphics[width=1.0\columnwidth]{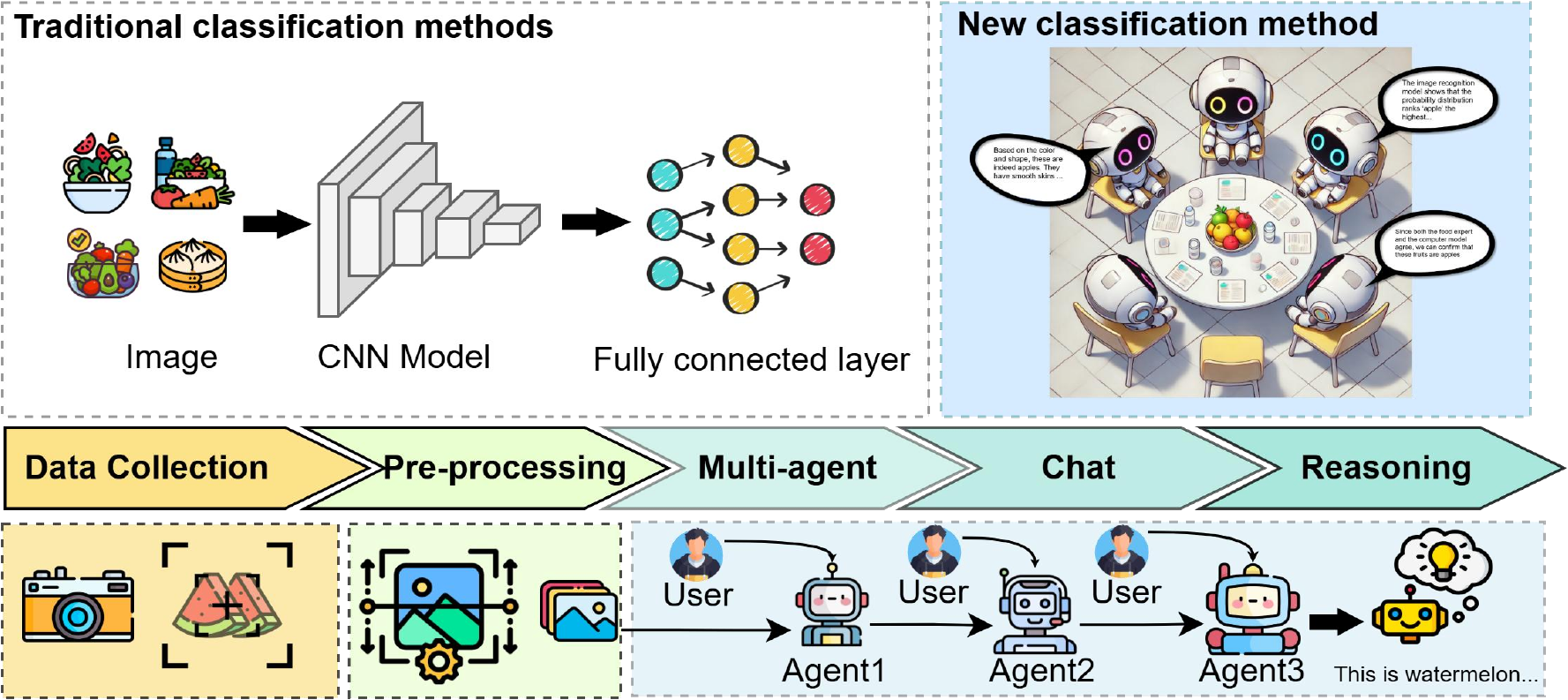}
	\caption{Overview of the proposed multi-agent, training-free classification framework. The top row contrasts traditional CNN-based classification with our multi-agent reasoning approach, where multiple specialized agents collaboratively analyze visual input. The bottom row illustrates the pipeline from data collection and pre-processing to multi-agent chat and reasoning, culminating in the final classification output.}
	\label{FIG:1}
\end{figure}

\section{Materials and Methods}
\subsection{Materials and Datasets}
\subsubsection{Food Datasets}
\begin{figure}[t]
    \centering
    \includegraphics[width=1.0\columnwidth]{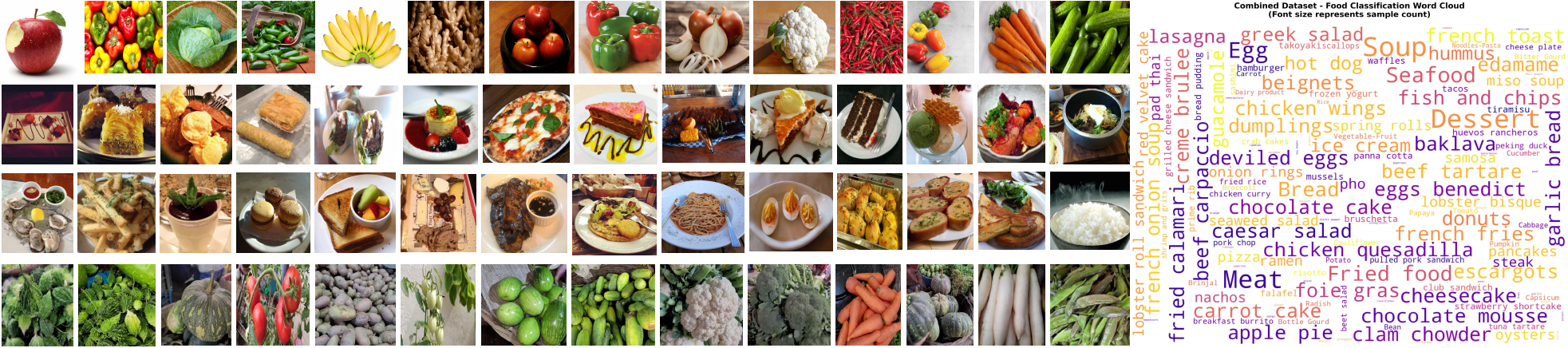}
    \caption{Examples of food image datasets used in this study: (a) Fruit-10, (b) Fruit and Vegetable Disease, (c) Food11, and (d) Food101.}
    \label{fig:datasets}
\end{figure}

To evaluate the proposed framework, four publicly available food image datasets were used, covering both fruit and vegetable classification and general food classification tasks. As shown in Figure~\ref{fig:datasets}.

\textbf{Fruit-10} contains 3,374 images across 10 fruit categories (e.g., apple, banana, cherry, mango), captured under diverse lighting and background conditions.
\footnote{\nolinkurl{https://www.kaggle.com/datasets/karimabdulnabi/fruit-classification10-class}}

\textbf{Fruit and Vegetable Disease (FVD)} comprises 30,000 images spanning 14 types of fruits and vegetables in both healthy and diseased states (e.g., fresh vs. rotten apples).\footnote{\nolinkurl{https://www.kaggle.com/datasets/muhammad0subhan/fruit-and-vegetable-disease-healthy-vs-rotten}}  

\textbf{Food11}, developed by the Multimedia Signal Processing Group at EPFL, includes 16,643 images across 11 broad food categories (e.g., bread, dairy, meat, vegetables). Images exhibit substantial variability in perspective, illumination, and background.  

\textbf{Food101} consists of 101,000 images across 101 categories, introduced by Bossard et al.~\cite{bossard2014food}. It combines high-quality and noisy images to reflect real-world complexity, making it suitable for testing robustness against label noise and presentation diversity.  

These datasets collectively cover fine-grained fruit recognition, freshness detection, and general dish classification, providing a comprehensive benchmark for evaluating food recognition models.

\subsubsection{Data processing}
In the field of deep learning, systematic data preprocessing is often necessary to effectively use the selected public food image datasets for model training and evaluation. The goal is to ensure that the input images meet the model's input specifications and enhance the diversity of the data. All images are first uniformly scaled to the model-specified resolution (e.g., 224×224, 336×336, or higher pixels). Pixel values are then normalized, usually mapping pixel values to the model's expected range based on the statistics used during model pretraining to ensure numerical stability. To improve the model's robustness to common variations in real food images, operations such as random horizontal flipping, random cropping, small-angle random rotation, and random brightness, contrast, and saturation adjustments are often applied. These operations simulate the natural variations that food may encounter during photography, storage, and display, and help the model learn more generalized food feature representations.

In contrast, this study exploits the core advantage of the visual language model (VLM), which is its ability to directly process raw image inputs and fully leverage the strong prior knowledge gained in large-scale multimodal pre-training. We directly input the standardized resized and normalized food images into the VLM. Thanks to its internal self-attention mechanism, VLM can dynamically focus on the most relevant areas and features in the image. This approach simplifies the input process and helps improve the versatility of food data input.

\subsubsection{Background and Motivation}

Traditional food image classification methods mainly rely on supervised learning, which uses a large amount of annotated data to train deep models to recognize predefined categories. Although such methods are effective in restricted scenarios, they face two core limitations: (1) the cost of collecting and annotating large-scale food image datasets is high, especially for fine-grained classification or specific regional cuisines, (2) their generalization ability is limited when encountering novel or ambiguous dishes not covered by the training set.

The progress of visual language models (VLMs) provides a promising alternative. These models have accumulated rich knowledge of visual concepts and semantic associations through large-scale image and text pre-training, and have strong prior capabilities. VLMs can understand and describe visual inputs in natural language, so that they can reason about images beyond a fixed set of labels. However, most existing VLM-based methods still rely on single-step reasoning, which limits the model's full reasoning potential. This motivates us to explore a multi-round conversational reasoning strategy that enables VLMs to collaboratively inspect images, propose hypotheses, ask questions for clarification, and gradually improve conclusions without relying on labeled training data.

\begin{figure}
	\centering
	\includegraphics[width=1.0\columnwidth]{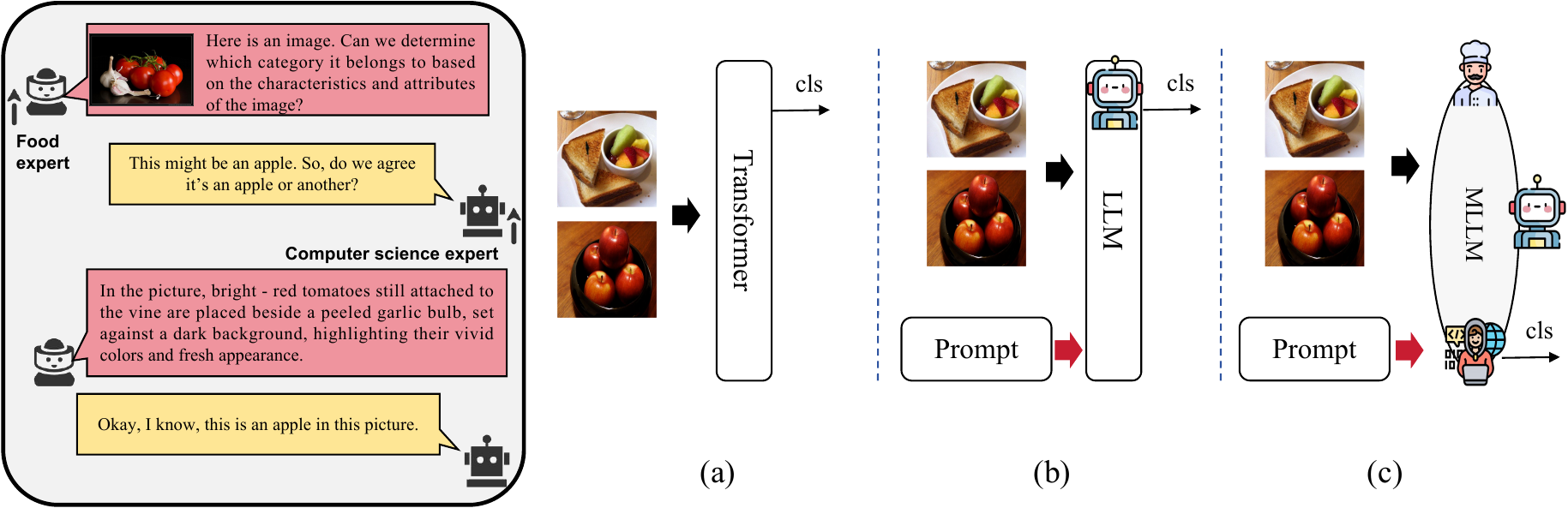}
	\caption{Overall framework of the proposed MultiFoodChat system. The model employs multi-turn dialogue between domain agents to improve food image classification accuracy, effectively handling fine-grained recognition tasks where visual-only models often fail.}
	\label{fig:dialog}
\end{figure}

\subsubsection{Visual-Language Model Architecture}
Our dialogue system is based on the Qwen3 Visual Language Model (VLM), a large-scale multimodal architecture that can understand visual input and generate natural language descriptions. Specifically, the model is built on a pre-trained multi-lingual language model (MLLM) and fine-tuned to learn to gain a more comprehensive understanding of food data. The visual module uses a food image $I \in \mathbb{R}^{W\times H\times C}$ as input and is processed by a visual encoder $f_v$ (based on the pre-trained ViT-L/14 model \cite{vit}) to extract feature representations $v=f_v(I)\in \mathbb{R}^d$, where $W$ and $H$ are the width and height of the image, respectively, and $C$ represents the number of channels.

Subsequently, the visual features $v$ are projected to the word embedding space of the language model through a linear layer with a trainable projection matrix to align the visual features with the text embedding space, resulting in the aligned visual feature embedding $H_v$. Meanwhile, the text module uses the prompt $X_q$ consisting of the task description, dialogue instructions, food description features, and food list as input.

The language model $\Phi(\cdot)$ generates output results based on the aligned visual features $H_v$ and the dynamic text feature sequence, expressed as:
\begin{equation}
T(y_n) = \Phi\left(H_v, \{H_{q}^{(t)}\}_{t=1}^n\right).
\end{equation}.

Where $n$ represents the current number of dialogue turns.

As shown in Figure~\ref{fig:dialog}, the fine-tuned model can classify food according to prompt instructions through images and dynamic dialogue information flow.

Existing deep learning methods for food image classification usually only support single image input. However, in actual application scenarios, food image data comes from various sources (such as user uploads, restaurant menus, and nutrition databases), and there are differences in image formats and content. Therefore, the visual processing module needs to have the ability to flexibly receive a single image as input. This design eliminates the need to redesign the core architecture of the system for food images from different sources or formats, ensuring the versatility and adaptability of the model. Specifically, our system design is optimized around a single food image input to meet the needs of the widest range of applications.

\subsection{Methods}
\label{sec:dialog_mode}

\begin{figure}[t]
    \centering
    \includegraphics[width=1.0\columnwidth]{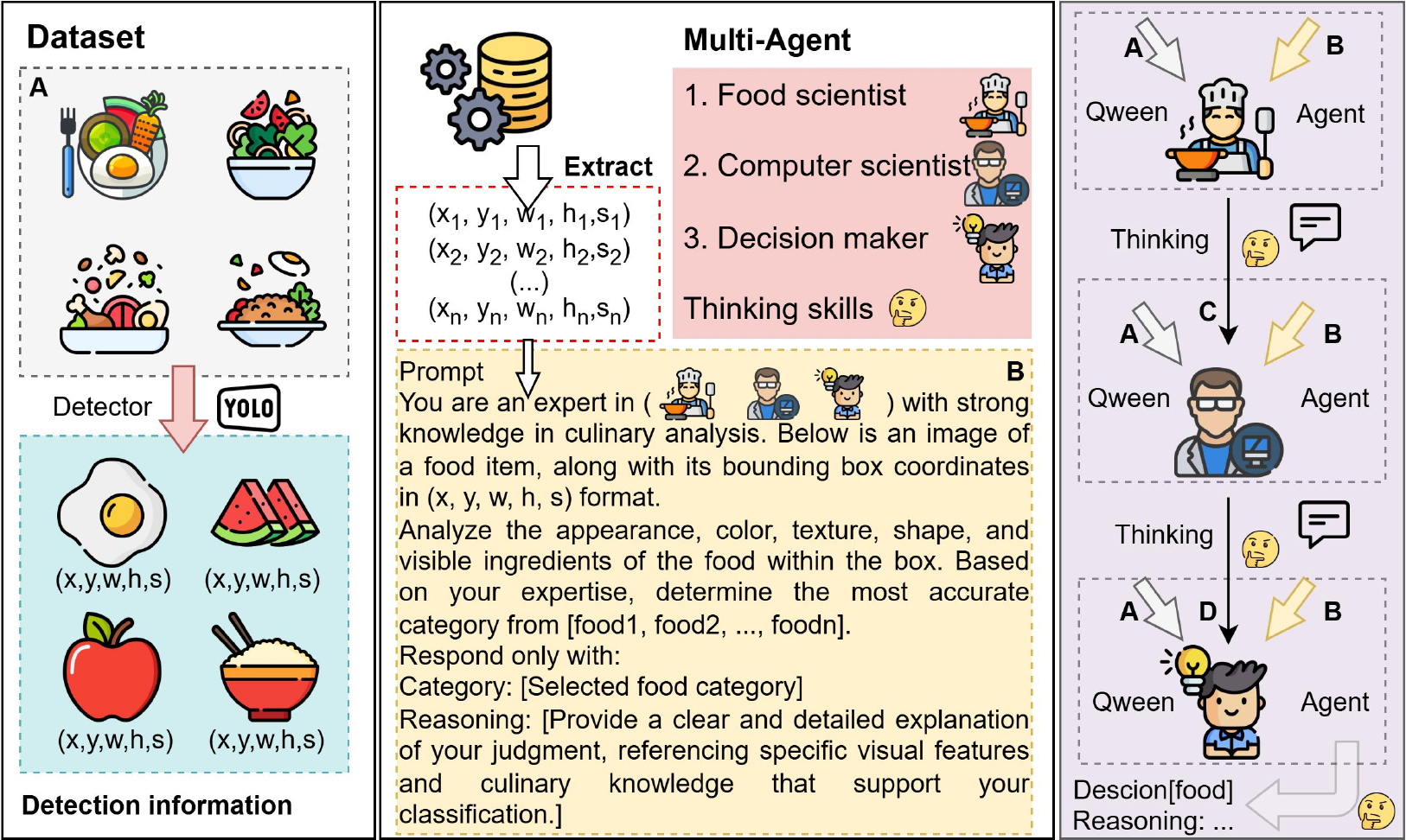}
    \caption{Multi-agent food classification framework. YOLO detects food items, and a team of agents (food scientist, computer scientist, decision maker) collaboratively reasons to generate the final category and explanation. Note: The prompt shown in the figure is a simplified example.}
    \label{fig:method}
\end{figure}

Our task is to use visual and conversational data for zero-shot object recognition. Specifically, our task requires combining visual and conversational data, where visual data provides clues for recognition and conversational patterns guide the recognition process. Ultimately, by analyzing the results of multiple rounds of conversation, we can identify objects that are difficult to distinguish based on visual information alone, as shown in Figure~\ref{fig:method}. Our approach consists of four tightly coupled modules: Object Perception Token Extraction, Visual Feature Encoding, Multi-turn Dialogue Mechanism, and Interactive Reasoning Agent with Prompt Engineering.

\subsubsection{Pipeline}

Each food image is represented by a ternary input $M=\{I, C, L\}$. These data are: (a) the input image $I$, which reflects the semantic information of the food in the RGB image, (b) the coordinate information $C$, which provides the 2D coordinate values of objects in the image, used to highlight the foreground information, (c) the conversation text $L$ through multiple rounds of text conversation, the LLM enhances its understanding and reasoning capabilities. Figure~~\ref{fig:method} shows the overall network structure. Our method is applicable to a dataset containing food images. Each food image is also associated with multiple sets of coordinate information $C=\{(x_1,y_1,w_1,h_1),(x_2,y_2,w_2,h_2),...,(x_n,y_n,w_n,h_n)\}$, where $(x,y)$ represents the coordinates of the center point of the object, and $(w,h)$ represents the height and width of the object in the image. Finally, we use $y \in \{1,2,3,4,...,k\}$ to represent the model output, i.e., the multi-classification result. Our goal is to learn a function that maps the input image data $I$, the coordinate information $C$, and the conversation feature $L$ to the output $y$, that is, satisfies the relationship $(I, C, L) \xrightarrow{} y $.

\subsubsection{Object Perception Token}

Before conducting multi-agent dialogue, it is necessary to accurately obtain the coordinate information $C$, which is a key step in achieving effective object perception tokens. To this end, we employ YOLOX~\cite{ge2021yolox}, a state-of-the-art real-time object detector. 

Given an input RGB image $I \in \mathbb{R}^{H \times W \times 3}$, YOLOX extracts multi-scale visual features through a backbone network $\mathcal{B}(\cdot)$ and aggregates them via a feature pyramid network (FPN). The detection head $\mathcal{D}(\cdot)$ outputs bounding boxes and class probabilities. Formally, the process can be expressed as:
\begin{equation}
F = \mathcal{B}(I), \qquad 
Z = \mathcal{D}(F),
\end{equation}
where $F$ denotes the multi-scale feature maps, and $Z$ represents the raw detection predictions. Each prediction $z_i \in Z$ consists of a bounding box and category scores:
\begin{equation}
z_i = \big[ (x_i, y_i, w_i, h_i), \; p_i \big],
\end{equation}
where $(x_i, y_i)$ is the center coordinate of the $i$-th box, $(w_i, h_i)$ are its width and height, and $p_i \in [0,1]^K$ is the confidence distribution over $K$ categories. 

To refine predictions, we apply non-maximum suppression (NMS) with threshold $\tau$:
\begin{equation}
C = \text{NMS}(Z, \tau) = \{ (x_j, y_j, w_j, h_j) \}_{j=1}^n,
\end{equation}
yielding the final set of $n$ high-confidence bounding boxes $C$. 

These coordinates not only highlight foreground regions of interest but also act as \textit{perception tokens} that are injected into subsequent multi-agent dialogue prompts. This ensures that all agents reason over localized and accurate visual information, improving both scene understanding and decision reliability.

\subsubsection{Multi-turn Dialogue Mechanism}

In our implementation, we chose Qwen3~\cite{qwen3} as the foundational LLM. This open-source multimodal model boasts powerful text understanding and visual perception capabilities, capable of processing both text and image inputs and supporting cross-modal information fusion and reasoning. In the conversational flow shown in Figure~\ref{fig:method}, the model, based on prompts, combines the visual features and coordinate information $C$ of the input image $I$ to gradually infer and output specific food categories through multiple rounds of interactive dialogue.

The MultiFoodChat system uses the pre-trained ViT as the visual encoder, responsible for encoding the input image into high-dimensional visual features. Subsequently, a linear layer with a trainable projection matrix maps the visual features $H_v$ to a dimension aligned with the text word embedding space. This ensures that the visual features $H_v$ and the text features $H_q$ have the same representation dimensionality in the same semantic space, thus achieving effective cross-modal fusion.

To construct the conversational prompt $L$, we designed a structured prompt template for the system, which includes clear task instructions, constraints, and prior knowledge. Specifically, we embed the object's coordinate information $C$ into the prompt. The coordinate information is used to enhance the model's focus on the image's foreground, improving object localization. At each turn \(t\in\{1,\dots,T\}\), the \textit{food scientist}, \textit{computer scientist}, and \textit{decision maker} produce outputs in a fixed order; the decision maker aggregates evidence and issues the final label \(y \in \mathcal{Y}\).

\subsubsection{Interactive Reasoning Agent}
\label{sec:multiagent}

We employ a multi-agent dialogue scheme to enable collaborative reasoning and classification on food images. The approach leverages the pretrained model's prior knowledge (food semantics, vision--language alignment, and natural-language reasoning) while assigning \emph{specialized roles} to instantiate complementary expertise. This human-like division of labor improves robustness and interpretability on ambiguous samples.

\paragraph{Roles and data flow.}
Let \(\mathcal{Y}\) be the set of valid food categories and \(\mathcal{R}\) the space of textual rationales. 
Given \(M=\{I,C,L\}\) with image \(I\), normalized boxes \(C=\{(x_i,y_i,w_i,h_i)\}_{i=1}^n\), and dialogue history \(L\), three agents interact in a fixed order:

\noindent\textbf{Food Scientist} (\(\mathrm{Agent}_{\text{food}}\)). A domain expert in food nutrition and taxonomy; it proposes a candidate class and a rationale using semantic priors and foreground cues:
\begin{equation}
\label{eq:agent_food}
\begin{aligned}
(\hat{y}_{\text{food}},\, r_{\text{food}}) 
  &= \mathrm{Agent}_{\text{food}}(I, C, L),
\hat{y}_{\text{food}} &\in \mathcal{Y}, \qquad r_{\text{food}} \in \mathcal{R}.
\end{aligned}
\end{equation}

\noindent\textbf{Vision Analyst} (\(\mathrm{Agent}_{\text{vision}}\)). A computer-vision specialist that verifies low-level evidence (texture, shape, color) and spatial plausibility, refining the hypothesis:
\begin{equation}
\label{eq:agent_vision}
\begin{aligned}
(\hat{y}_{\text{vision}},\, r_{\text{vision}})
  &= \mathrm{Agent}_{\text{vision}}(I, C, L, \hat{y}_{\text{food}}, r_{\text{food}}),
\hat{y}_{\text{vision}} &\in \mathcal{Y}, \qquad r_{\text{vision}} \in \mathcal{R}.
\end{aligned}
\end{equation}

\noindent\textbf{Decision Maker} (\(\mathrm{Agent}_{\text{decider}}\)). A comprehensive arbiter that synthesizes both perspectives with the original inputs to produce the final label:
\begin{equation}
\label{eq:agent_decider}
\begin{aligned}
y \;=\;&\ \mathrm{Agent}_{\text{decider}}(I, C, L, \hat{y}_{\text{food}}, r_{\text{food}},\ \hat{y}_{\text{vision}}, r_{\text{vision}}), \qquad y \in \mathcal{Y}.
\end{aligned}
\end{equation}

\noindent
Here, each \(r_{\star}\) is a textual explanation supporting the corresponding hypothesis and is appended to \(L\) for subsequent turns.

For reproducibility, we use concise role prompts defining responsibilities and output format (``\texttt{Category: ...; Reasoning: ...}''). The Food Scientist must ground claims in visible cues inside the box; the Vision Analyst must explicitly \emph{agree/disagree/refine} the prior judgment with cited visual evidence; the Decision Maker provides a short synthesis and the final category \(y\). 

\section{Experiments and Analysis}
\subsection{Experimental Setup}
\label{sec:setup-keypoints}

\begin{itemize}
  \item \textbf{Evaluation protocol.} We directly evaluate the model on the test splits of four datasets. All experiments were conducted with Python 3.9 and PyTorch 2.0.1 under CUDA 11.1. The operating system was Ubuntu 22.04 LTS, and all computations were performed on an NVIDIA A100 GPU. The decoding parameters for Qwen3 were set to a temperature of 0.2 and a maximum of 512 new tokens.

  \item \textbf{Detection settings.} Object-perception tokens (OPT) were generated using YOLOX-M (v0.3.0) implemented in PyTorch. Input images were resized to \textbf{640$\times$640} before inference. The confidence (score) threshold was set to 0.5, and non-maximum suppression (NMS) was applied with an IoU threshold of 0.5. The maximum number of detections per image was 20, and inference was performed with a batch size of~16 in FP16 precision. All other hyperparameters followed the default YOLOX-M configuration.
\end{itemize}

\begin{itemize}
  \item \textbf{Metrics.} We report overall accuracy (ACC), recall, and F1. Let $TP$, $FP$, $FN$ be true positives, false positives, and false negatives (computed per class); macro-averaged scores are used unless noted otherwise:
  \begin{align}
    \mathrm{ACC} &= \frac{\sum_{k=1}^{K} TP_k}{\sum_{k=1}^{K} (TP_k+FP_k+FN_k)}, \\
    \mathrm{Recall}_k &= \frac{TP_k}{TP_k+FN_k}, \\
    \mathrm{F1}_k &= \frac{2\,TP_k}{2\,TP_k+FP_k+FN_k}, \\
    \mathrm{Recall} &= \frac{1}{K}\sum_{k=1}^{K}\mathrm{Recall}_k, \quad
    \mathrm{F1} = \frac{1}{K}\sum_{k=1}^{K}\mathrm{F1}_k. 
  \end{align}
\end{itemize}

\subsection{Comparative Evaluation on Benchmark Datasets}

Our MultiFoodChat model was compared with models such as VGG16~\cite{vgg}, ResNet18/50~\cite{resnet}, MobileNet~\cite{mobilenets}, MobileNetV2~\cite{mobilenetv2}, and EfficientNet~\cite{efficientnet} on the Fruit-10 and Fruit and Vegetable Disease (FVD) datasets. The results are shown in Table 1. On the Fruit-10 dataset, MobileNetV2 achieved the highest accuracy of 95.22\%, followed by MobileNet at 92.10\%. MultiFoodChat achieved 90.19\%. While this still fell short of the state-of-the-art result by approximately 5\%, it significantly outperformed VGG16 (85.63\%), ResNet18 (86.72\%), and EfficientNet (89.66\%), achieving performance very close to the state-of-the-art model. On the FVD dataset, MobileNetV2 still performed best with an accuracy of 95.93\%. MultiFoodChat achieved 91.88\%, about 4\% lower than the highest value, but still surpassed VGG16 (90.93\%), ResNet18 (90.61\%), and EfficientNet (92.07\%). This shows that even with large-scale fruit and vegetable data, MultiFoodChat can approach the best performance.

\begin{table*}[htbp]
\centering
\begin{subtable}[t]{0.48\textwidth}
\centering
\caption{Fruit-10 classification dataset}
\begin{tabular}{lccc}
\toprule
Model       & Acc    & Recall & F1    \\
\midrule
vgg16       & 85.63  & 84.30  & 83.90 \\
resnet18    & 86.72  & 85.40  & 85.00 \\
resnet50    & 87.80  & 86.50  & 86.10 \\
mobilenet   & 92.10  & 91.03  & 90.60 \\
mobilenetv2 & \textcolor{red}{95.22}  & \textcolor{red}{94.14}  & \textcolor{red}{93.67} \\
efficientnet& 89.66  & 88.40  & 88.09 \\
Ours         & \textcolor{blue}{90.19} & \textcolor{blue}{88.80} & \textcolor{blue}{88.38} \\
\bottomrule
\end{tabular}
\end{subtable}
\hfill
\begin{subtable}[t]{0.48\textwidth}
\centering
\caption{Fruit and Vegetable Disease dataset}
\begin{tabular}{lccc}
\toprule
Model       & Acc    & Recall & F1    \\
\midrule
vgg16       & 90.93  & 85.19  & 85.10 \\
resnet18    & 90.61  & 85.64  & 84.83 \\
resnet50    & 92.39  & 87.41  & 86.46 \\
mobilenet   & 92.72  & 87.77  & 86.09 \\
mobilenetv2 & \textcolor{red}{95.93}  & \textcolor{red}{90.84}  & \textcolor{red}{90.03} \\
efficientnet& 92.07  & 87.55  & 86.52 \\
Ours         & \textcolor{blue}{91.88} & \textcolor{blue}{86.90} & \textcolor{blue}{86.03} \\
\bottomrule
\end{tabular}
\end{subtable}
\label{tab1}
\caption{Comparison of classification performance on Fruit and Vegetable datasets. The highest values are marked in \textcolor{red}{red}, and the second-highest in \textcolor{blue}{blue}.}
\end{table*}


On two larger and more challenging food image datasets, Food11 and Food101, our model, MultiFoodChat, was compared with AlexNet~\cite{alexnet}, VGG16, ResNet50/152, InceptionV3~\cite{InceptionV3}, DenseNet161~\cite{densely}, RexNet~\cite{han2021rethinking}, and the improved methods ASTFF~\cite{ASTFF} and GCAM~\cite{GCAM}. The results are shown in Table 2. On Food11, ASTFF achieved a top-tier accuracy of 95.04\%, while MultiFoodChat achieved 93.53\%, only about 1.5\% lower than the state-of-the-art result. This performance also outperformed commonly used models such as ResNet50 (90.32\%) and DenseNet161 (93.06\%). On Food101, ASTFF still performed best, reaching 93.06\%, while MultiFoodChat's accuracy was 87.70\%, about 5\% lower than the highest result, but it had a significant advantage over mainstream models such as VGG16 (79.02\%) and ResNet50 (85.65\%).


\begin{table*}[htbp]
\centering
\begin{tabular}{lcccccc}
\toprule
\multirow{2}{*}{Model} & \multicolumn{3}{c}{Food11} & \multicolumn{3}{c}{Food101} \\
\cmidrule(lr){2-4} \cmidrule(lr){5-7}
 & Acc & Recall & F1 & Acc & Recall & F1 \\
\midrule
AlexNet     & 82.07 & 77.65 & 76.92 & 55.89 & 51.34 & 50.63 \\
vgg16       & 87.17 & 82.64 & 81.92 & 79.02 & 74.38 & 73.65 \\
resnet50    & 90.32 & 85.71 & 84.96 & 85.65 & 81.06 & 80.21 \\
resnet152   & 91.34 & 86.72 & 85.97 & 86.61 & 82.03 & 81.28 \\
InceptionV3 & 89.06 & 84.43 & 83.72 & 84.15 & 79.62 & 78.87 \\
densenet161 & \textcolor{blue}{93.06} & \textcolor{blue}{88.52} & \textcolor{blue}{87.73} & \textcolor{blue}{86.94} & \textcolor{blue}{82.37} & \textcolor{blue}{81.59} \\
RexNet      & 93.47 & 88.91 & 88.15 & 85.59 & 81.08 & 80.27 \\
ASTFF       & \textcolor{red}{95.04} & \textcolor{red}{90.41} & \textcolor{red}{89.62} & \textcolor{red}{93.06} & \textcolor{red}{88.52} & \textcolor{red}{87.69} \\
GCAM        & 94.32 & 89.73 & 88.95 & 91.11 & 86.42 & 85.67 \\
Ours        & 93.53 & \textcolor{blue}{93.02} & \textcolor{blue}{92.39} & 87.70 & \textcolor{blue}{85.62} & \textcolor{blue}{85.47} \\
\bottomrule
\end{tabular}
\caption{Comparison of classification performance on Food11 and Food101 datasets. The highest values are marked in \textcolor{red}{red}, and the second-highest in \textcolor{blue}{blue}.}
\label{tab2}
\end{table*}

It's worth noting that MultiFoodChat is a training-free model, meaning it can be used directly without additional training. Comparing it to CNN and Transformer models requires training with large-scale labeled data. In comparisons with unsupervised methods SimCLR~\cite{chen2020simple}, SwAV~\cite{caron2020unsupervised}, BYOL~\cite{grill2020bootstrap}, SimSiam~\cite{chen2021exploring}, MoCov2~\cite{chen2020improved}, and DINO~\cite{caron2021emerging}, MultiFoodChat demonstrates significant advantages (see Table 3). For example, DINO achieves a peak accuracy of only 61.40\% in unsupervised scenarios, while MultiFoodChat achieves 87.70\%, a performance improvement of over 25 percentage points. This demonstrates that, even without additional training, MultiFoodChat outperforms most traditional supervised models in food image classification tasks and significantly surpasses existing unsupervised learning methods, demonstrating its strong versatility and applicability. Furthermore, leveraging the prior knowledge embedded in a large-scale language model, MultiFoodChat effectively integrates visual and semantic information without training, achieving performance approaching or even exceeding that of some supervised models.

\begin{table}[htbp]
\centering
\begin{tabular}{lc}
\toprule
Model   & Acc  \\
\midrule
SimCLR  & 51.00 \\
SwAV    & 54.70 \\
BYOL    & 47.70 \\
SimSiam & 44.50 \\
MoCov2  & 53.90 \\
DINO    & 61.40 \\
Ours     & 87.70 \\
\bottomrule
\end{tabular}
\label{tab3}
\caption{Comparison of unsupervised learning models}
\end{table}

\subsection{Ablation Study of Model Components}

To verify the contribution of each module to overall performance, we conducted progressive ablation experiments on four datasets: Fruit-10, Fruit and Vegetable Disease (FVD), Food11, and Food101. The results are shown in Table 4. (a) directly inputs images and simple prompt words into the model for prediction; (b) introduces OPT (Object Perception Token) on this basis; (c) further incorporates multi-turn reasoning; and (d) interactive reasoning multi-Agent~(IRA). It can be observed that with the gradual introduction of modules, the model's classification performance shows a continuous improvement.

First, the introduction of OPT enables the model to focus on the foreground during reasoning, avoiding interference from complex backgrounds. Without OPT, the model struggles to accurately capture key food features. However, with OPT added, accuracy significantly improved on both the Fruit-10 and FVD datasets, demonstrating the importance of explicit cues for the target region in food classification.

Second, the multi-turn reasoning mechanism allows the model to continuously refine its initial predictions during multi-step conversations, gradually improving the stability and reliability of the results. Comparing the base model with the version incorporating multi-turn reasoning, the accuracy on both Food11 and Food101 improved by approximately 2–3 percentage points, demonstrating that iterative reasoning effectively mitigates the uncertainty of single-step predictions, particularly in complex food tasks with subtle inter-class differences.

Finally, IRA improves overall robustness and interpretability by introducing different "scientist" roles to perform reasoning at the semantic, visual, and comprehensive decision-making levels of food. Compared with single-agent models, IRA achieves particularly significant performance improvements on the Food101 dataset, bringing its accuracy close to or even exceeding that of some supervised methods, demonstrating the critical role of multi-view reasoning in complex food classification scenarios.

\begin{figure}
	\centering
	\includegraphics[width=0.8\columnwidth]{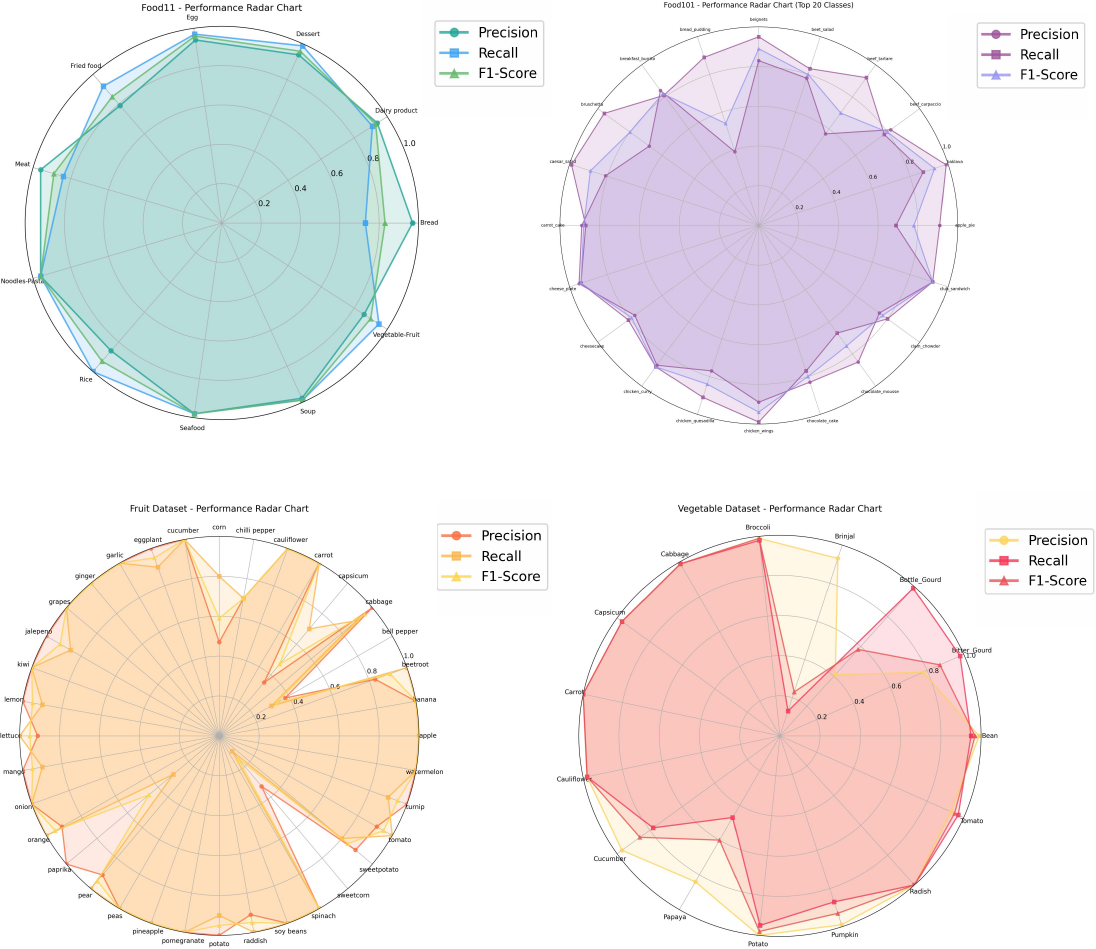}
	\caption{Radar charts showing class-wise Precision, Recall, and F1-score of MultiFoodChat across four benchmark food datasets.}
	\label{f5}
\end{figure}

\begin{figure}
	\centering
	\includegraphics[width=0.8\columnwidth]{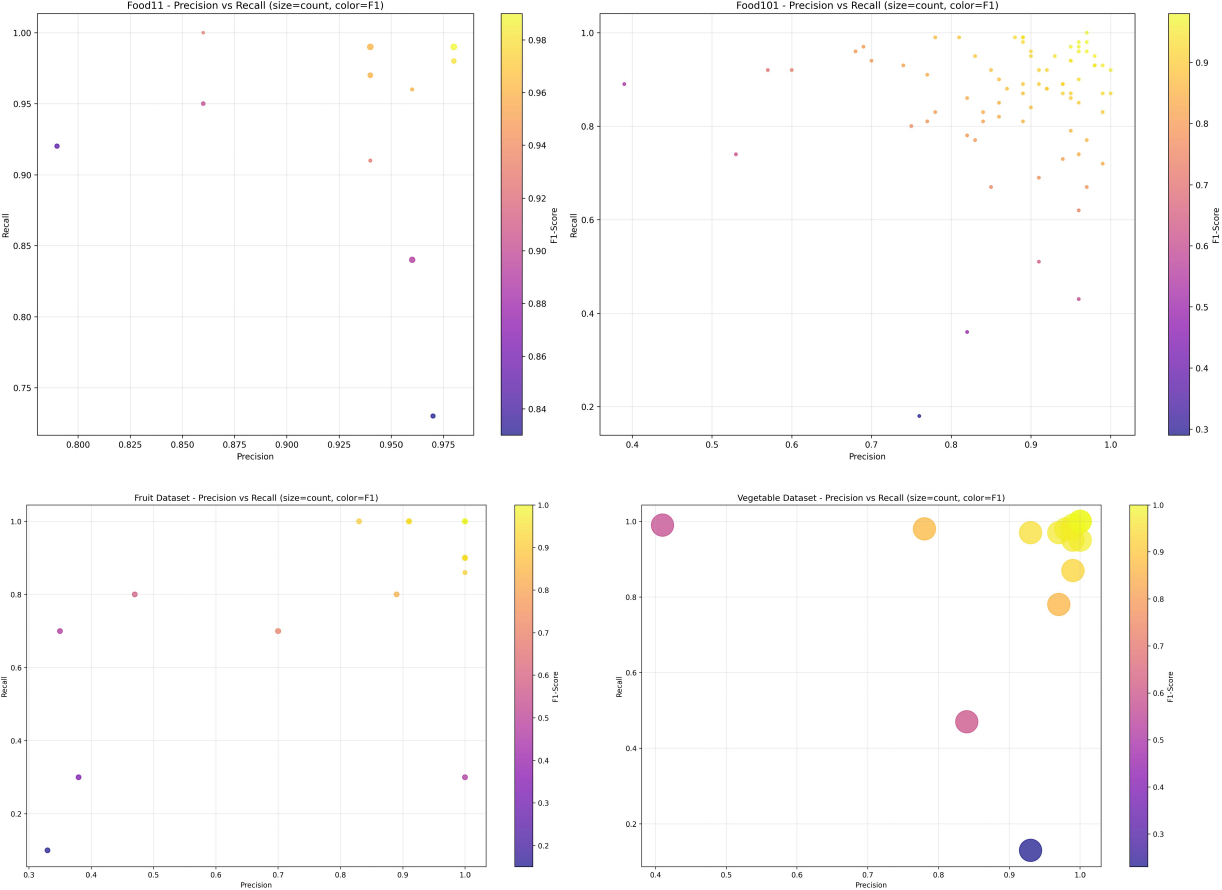}
	\caption{Precision–Recall scatter plots for MultiFoodChat across food datasets, with bubble size indicating sample count and color representing F1-score.}
	\label{f6}
\end{figure}

\begin{table*}[htbp]
\centering
\begin{tabular}{lccccccc}
\toprule
Setting & OPT & Multi-turn & IRA & Fruit-10 & FVD & Food11 & Food101 \\
\midrule
a    &  &            &             & 82.73 & 83.02 & 85.12 & 83.45 \\
b    & \checkmark &  &             & 85.45 & 84.77 & 87.74 & 85.12 \\
c    & \checkmark & \checkmark &   & 89.73 & 90.92 & 92.63 & 86.43 \\
d & \checkmark & \checkmark & \checkmark  & 90.19 & 91.88 & 93.53 & 87.70 \\
\bottomrule
\end{tabular}
\caption{Ablation study of different components on multiple datasets. 
OPT: Object Perception Token. IRA: Interactive Reasoning Multi-Agent. FVD: Fruit
 and Vegetable Disease dataset}
\label{tab4}
\end{table*}

Overall, the three modules each make significant contributions. OPT provides stable foreground perception, multi-round reasoning enhances iterative decision correction, and multi-agent collaboration further ensures the robustness and interpretability of classification results. The combination of these three components enables the model to achieve optimal performance on all four datasets, fully demonstrating the rationality and effectiveness of the design.

\subsection{Visualization and Performance Analysis}
\begin{figure}
	\centering
	\includegraphics[width=0.8\columnwidth]{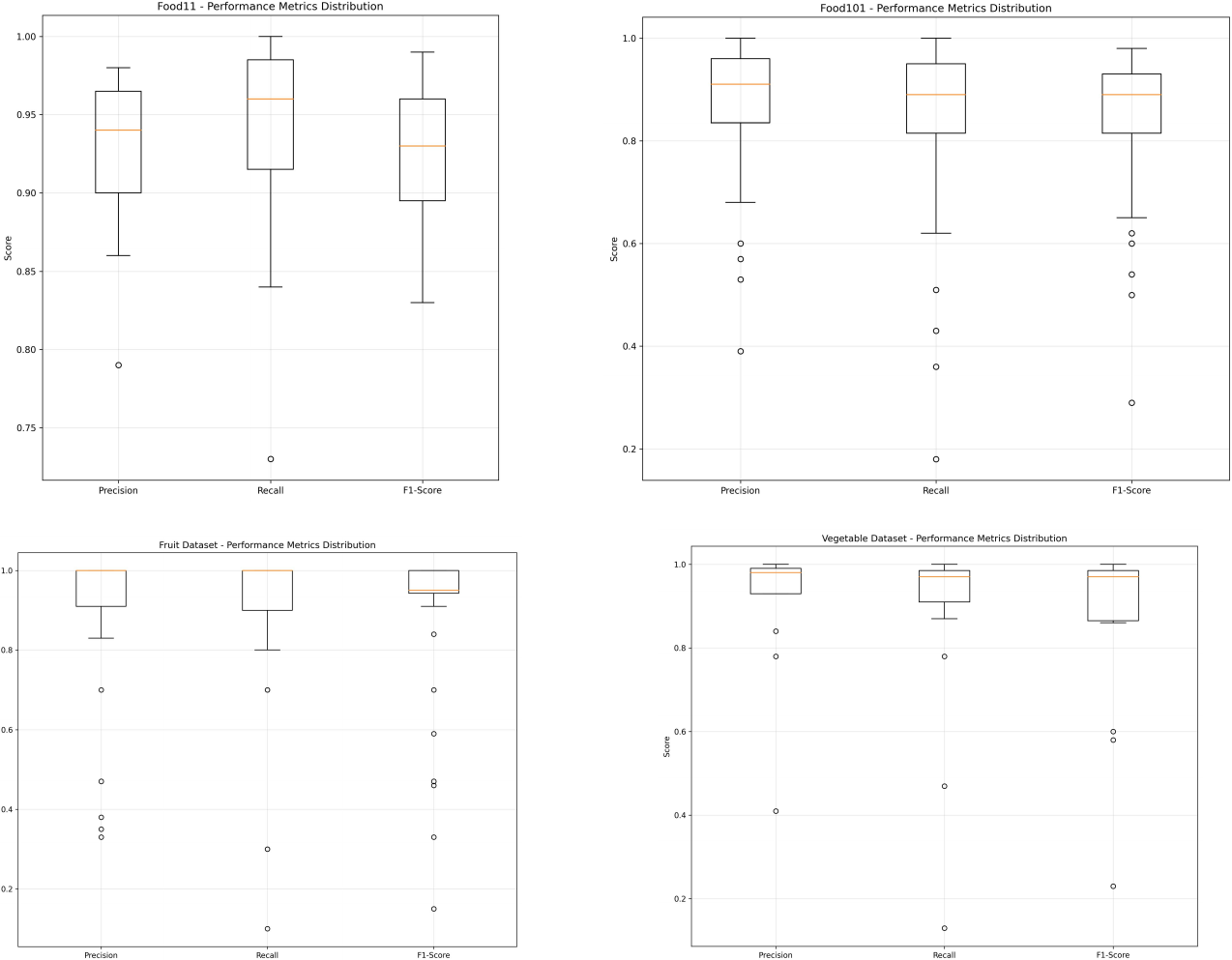}
	\caption{Distribution of Precision, Recall, and F1-score for MultiFoodChat across different food datasets, illustrating performance consistency and variability.}
	\label{f7}
\end{figure}

To more comprehensively evaluate the model's performance across different datasets and categories, we plotted radar charts, precision-recall scatter plots, and performance distribution boxplots (see Figs.~\ref{f5}, \ref{f6}, and \ref{f7}). These visualizations illustrate the differences in model performance across the three metrics of Precision, Recall, and F1-score from different perspectives.

The radar chart shows that on the Food11 dataset, the model's overall performance is relatively balanced, with Precision, Recall, and F1-score remaining above 0.9 for almost all categories, demonstrating that the model maintains stable discrimination across most food categories. On Food101, while the overall trend remains positive, individual categories (such as visually similar desserts and beverages) experience a decrease in Recall, reflecting challenges faced by the model in scenarios with a large number of categories and subtle differences. For the Fruit-10 and FVD datasets, the radar charts also demonstrate that the model maintains high stability for most categories, but for categories with fewer samples, Recall and F1-score fluctuate slightly relative to Precision. The scatter plot further reveals the relationship between Precision and Recall. In the Food11 and FVD datasets, the majority of points are concentrated in the upper right region, with both Precision and Recall above 0.85, indicating that the model achieves both high precision and recall. In the Food101 dataset, while the majority of points remain in the high-precision range, a small number of categories fall into the relatively low-recall range, resulting in a stretched overall performance distribution and highlighting the imbalances inherent in complex, large-scale data. In the Fruit-10 dataset, the points are more concentrated, and the F1-score is generally warmer, indicating that the model maintains good performance even on small-scale, refined tasks.

The boxplots illustrate the statistical distribution characteristics of Precision, Recall, and F1-score. In the Food11 and Fruit-10 datasets, the medians of all three metrics were close to 0.95, with a small interquartile range, indicating stable and reliable classification results for most categories. In the Food101 and FVD datasets, while the overall medians remained high, some outliers were observed, indicating suboptimal performance in some difficult-to-classify categories. Notably, these outliers were mostly concentrated in categories with insufficient sample size or highly similar visual features, suggesting that future optimization efforts could focus on balanced category sampling and feature enhancement modeling.
Overall, the results show that our method achieves stable and balanced performance across most food categories, maintaining high levels of precision, recall, and F1-score. However, there is still room for improvement on the large-scale, fine-grained Food101 dataset. 

\section{Conclusion}
We presented \textbf{MultiFoodChat}, a food image classification framework that combines visual--linguistic reasoning with a multi-agent collaboration scheme. We evaluated the method on four benchmarks—Fruit-10, Fruit and Vegetable Disease, Food11, and Food101—and reported class-wise Precision, Recall, and F1. The results show that MultiFoodChat delivers strong and balanced performance across datasets without task-specific training. Ablation studies confirm the contribution of the Object Perception Token (OPT), multi-turn dialogue, and the multi-agent design. Complementary visual analyses (radar, PR scatter, and score distributions) indicate consistent behavior across most categories, supporting the method's effectiveness and robustness.

\section{Limitations and Future Work}
This study mainly validated the proposed reasoning framework using the Qwen3 series of large language models. Further comparative experiments with other mainstream models—such as ChatGPT, Gemini, and DeepSeek have not yet been performed. Expanding such comparisons in future work would help better understand the framework’s adaptability and consistency across different model architectures.

At present, our research represents an early exploration focused on food image classification and reasoning. In future studies, we plan to gradually extend the framework toward more comprehensive food analysis, such as exploring connections with chemical composition or nutritional information. Incorporating these aspects, together with larger and more diverse datasets and external knowledge sources (e.g., ingredient or recipe databases), may further enhance the model’s interpretability and practical relevance in real-world food applications.



\bibliography{ref}
\end{document}